# Sub-Pixel Accuracy Edge Fitting by Means of B-Spline

Ricardo Lucas Bastos Breder [1], Vania Vieira Estrela [*2], Joaquim Teixeira de Assis [3]

*State University of Rio de Janeiro (UERJ), Polytechnic Institute of Rio de Janeiro (IPRJ), CP 972825, CEP 28601-970, Nova Friburgo, RJ, Brazil*

[1] `r1c4rd0.luc45@gmail.com`
[2] `vestrela@iprj.uerj.br`
[3] `joaquim@iprj.uerj.br`

*Abstract*— Local perturbations around contours strongly disturb the final result of computer vision tasks. It is common to introduce *a priori* information in the estimation process. Improvement can be achieved *via* a deformable model such as the snake model. In recent works, the deformable contour is modeled by means of B-spline snakes which allows local control, concise representation, and the use of fewer parameters. The estimation of the sub-pixel edges using a global B-spline model relies on the contour global determination according to a Maximum Likelihood framework and using the observed data likelihood. This procedure guarantees that the noisiest data will be filtered out. The data likelihood is computed as a consequence of the observation model which includes both orientation and position information. Comparative experiments of this algorithm and the classical spline interpolation have shown that the proposed algorithm outperforms the classical approach for Gaussian and Salt & Pepper noise.

## I. INTRODUCTION

One of the greatest challenges of modeling remotely sensed images is the mixed pixel problem, that is, the level of spatial detail captured is less than the amount of expected details. This sub-pixel heterogeneity is important, but not readily capable of being known. In a traditional manner, each pixel is classified into one of many land cover types (hard classification), implying that land cover exactly fits within the bounds of one or multiple pixels. However, several pixels consist of a mixture of different classes. The solution to the mixed pixel problem typically centers on soft classification, which allows proportions of each pixel to be partitioned between classes. An important edge extraction approach is concerned with improving the detection accuracy and different sub-pixel edge detectors [10].

Popular approaches used to compute sub-pixel location of edges are based on the image moments [6, 11, 14, 15, 16, 18]. Image interpolation is another way to find sub-pixel edge coordinates. [13] presents a technique where the first derivative perpendicular to the edge orientation is approximated by a normal function. The edge location is this function maximum. Other works are based on linear [7], quadratic interpolation [1], B-spline [17], non-linear functions [8], and as a segmentation by-product [22, 23].

In all these methods, the estimation is local and does not include a noise model. Hence, the local perturbations heavily disturb the final result. A usual filtering approach requires introducing *a priori* information in the estimation process such as the snake model suggested by [9]. In newer works, the deformable contour is modeled using piecewise polynomial functions (B-spline snakes) [4, 12] allowing local control, concise representation, and it employs few parameters.

This work introduces a model for estimating sub-pixel edges using a global procedure based on a B-spline model [3, 18], but the statistical properties of the observations are computed and used in a Maximum Likelihood (ML) estimation framework which insures an efficient filtering of the noisy data. Section II briefly describes the classical B-spline formulation. Section III introduces the proposed extension to the sub-pixel case. Section IV discusses experimental results of this algorithm. Finally, Section V presents some conclusions.

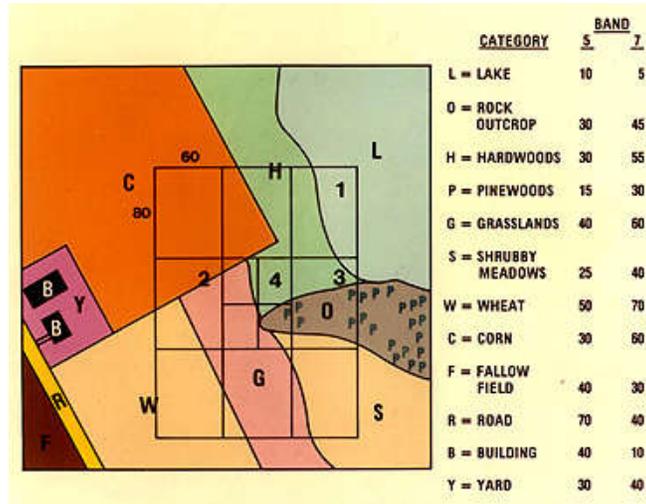

Fig. 1 – Illustration of the mixed-pixel problem [20].

## II. B-Spline Contour Formulation

Let $\{t_0, t_1, ..., t_{k-1}\}$ be the set of so-called knots. Spline functions are polynomial inside each interval $[t_{i-1}, t_i]$ and exhibit a certain degree of continuity at the knots (see [3]). The set $\{B_{i,n(t)}, i = 0, ..., k-n-1\}$ of the B-splines, constitutes a basis for the linear space of all the splines contained in $[t_n, t_{k-n}]$. Hence, a spline curve $f(t)$ of degree $n$ is given by:

$$f(t) = \sum_{i=0}^{k-n-1} p_i B_{i,n}(t),$$

where $p_i$ is the weight applied to the corresponding basis functions $B_{i,n}$. The B-Spline functions are defined by the recursion formulas in [2]:

$$B_{i,0}(t) = \begin{cases} 1 & if \\ 0 & else \end{cases} \quad t_i \leq t \leq t_{i+1}, \text{ and}$$

$$B_{i,n}(t) = \frac{t - t_i}{t_{i+n} - t_i} B_{i,n-1}(t) + \frac{t_{i+n+1} - t}{t_{i+n+1} - t_{i+1}} B_{i+1,n-1}(t).$$

The $B_{i,n(t)}$ are nonnegative, $J_i = (G_{yi} - G_{xi})$ and obey the relationship: $\sum_i B_{i,n(t)} = 1$ for all values of $t$. Without loss of generality, the index $n$ for the cubic B-spline can be dropped. Eq. (1) can be re-written in matrix notation as:

$$f(t) = B^T(t).\theta.$$

$v(t) = (x(t), y(t))$ is the vector of B-spline functions $B(t) = (B_0(t), B_2(t), ..., B_{N-1}(t))$ and $\theta$ is the weight vector: $\theta = (p_0, ..., p_{N-1})^T$. The 2D version of Eq. (1) describes an open parametric snake on the plane: $v(t) = (x(t), y(t))$. The pixels $p_i = (p_{xi}, p_{yi})$ are now 2D vectors and are called the control points.

## III. Subpixel B-Spline Fitting

This sub-section aims to obtain a likelihood expression for the observation, supposing additive Gaussian noise $b$ with standard deviation $\sigma_b$.

Let $v_e = \{v_e(i.h) = (x_e(i.h), y_e(i.h)), i \in \{1, ..., M\}\}$ be a set of edge pixels with integer coordinates $h$ computed by a classical algorithm. To simplify notation, $v_e(i.h) = v_e$ for all variables related to the pixel $i$.

For each pixel *i* belonging to this set, an observation vector $O_i = (X_i, H_i)$ is assumed, where vector $H_i = (H_{xi}, H_{yi})$ is the local gradient of the image and $X_i$ corresponds to a sub-pixel position estimation of the edge along $H_i$. $X_i$ is estimated in the pixel local coordinate system by means of a quadratic interpolation [5]. These variables are supposed to be the observation version of the true variables $(Y_i, G_i)$, where $(x_{oi}, y_{oi})$ are the cartesian coordinates of the observation $O_i$.

$v_s(t) = (x_s(t) = B(t).\theta_x, y_s(t) = B(t).\theta_y)$ is the edge B-spline model, where $\theta = (\theta_x, \theta_y)$ is the vector of *k* control points $(M \gg k)$. The likelihood becomes

$$P(O_i/\theta) = P(X_i/H_i,\theta) P(H_i/\theta).$$

To derive $P(X_i/H_i,\theta)$, these assumptions are made: $X_i$ depends only on $v_s(t(i)) = (X_s(t(i))) = x_{si}, y_s(t(i)) = y_{si}$, the corresponding edge point in the direction $H_i$ has polar coordinates $(Y_i, H_i)$ and $P(X_i/H_i, Y_i)$ is Gaussian. Then,

$$P(X_i/H_i,\theta) = P(X_i/H_i,Y_i) = \frac{1}{\sqrt{2\pi}\sigma_{Xi}} \exp\left(-\frac{[(X_{oi}-X_{si})+(Y_{oi}-Y_{si})]^2}{2\sigma_{Xi}^2}\right)$$

The computation of $\sigma_{Xi}$ is based on a modified version of the method described in [3]. Now, $P(H_i/\theta)$ requires the components of $H_i$. They can be computed by convolving the original image with the derivative kernels $K_x$ and $K_y$. Assuming an additive Gaussian noise, it is straightforward to show that $P(H_i/G_i)$ is also Gaussian. The cross-correlation and auto-correlation functions of the vertical and horizontal components of *H* are given by:

$$C_{xy}(x_i-x_j, y_i-y_j) = E[H_{xi}.H_{yj}] - E[H_{xi}]E[H_{yj}]$$
$$= \sigma_b^2 \delta(x_i-x_j, y_i-y_j) * K_x(x,y) * K_y^*(-x,-y).$$

$K_x^*$ and $K_y^*$ are the complex conjugate of the matrices of $K_x$ and $K_y$, respectively. This relation leads to

$$E[H_{xi}.H_{yi}/G_{xi}G_{yi}] - E[H_{xi}/G_{xi}G_{yi}]E[H_{yi}/G_{xi}G_{yi}] = 0,$$

and

$$E[H_{xi}.H_{xi}/G_{xi}] - E[H_{xi}/G_{xi}]E[H_{xi}/G_{xi}] = \sigma_b^2 \sum_{i,j} K_x(i,j)^2$$
$$= \sigma_b^2 \sum_{i,j} K_y(i,j)^2 = \sigma_H^2.$$

$P(H_i/\theta)$ represents the conditional expected value of *w* assuming *z*. Let $J_i$ be the vector defined by $J_i = (G_{yi}, -G_{xi})$ and let $R_i$ be the unit vector collinear to $v_s$ in $v_{si}$. $H_i$ and $J_i$ are clearly related by a normal distribution. To find $P(H_i/\theta)$, the well known property of the edge being orthogonal to the local gradient is used:

$$P(H_i/\theta) = P(H_i/R_i(\theta)) = \int \frac{P(H_i/J_i)P(J_i)}{P(R_i(\theta))},$$

for all the values of $J_i$ collinear to $R_i$, that is $J_i = a\, R_i$. It can be rewritten as:

$$P(O/v_s) = \prod_i P(O_i/v_s).$$

The gradient does not have a privileged direction. Uniform regions are supposed to be more frequent than the regions with a high gradient magnitude. Thus, the *a priori* probability of $J_i$ can be modelled by

$$P(J_i) = \frac{1}{\sqrt{\pi}\Sigma} \exp\left(-\frac{||J_i||^2}{\Sigma^2}\right) = \frac{1}{\sqrt{\pi}\Sigma} \exp\left(-\frac{a^2}{\Sigma^2}\right).$$

Combining the two previous expressions results in

$$P(H_i / \theta) = K.\exp\left(-\frac{H_{yi}^2 + H_{xi}^2}{2\sigma_H^2 + \Sigma^2}\right) \times$$

$$\exp\left(-\frac{(H_{xi}R_{xi} + H_{yi}R_{yi})^2 \Sigma^2}{2\sigma_H^2(2\sigma_H^2 + \Sigma^2)}\right).$$

$K$ is a normalization constant involving $P(R_i)$, $\Sigma$ and $\sigma H$. If $\Sigma \ggg \sigma H$, then:

$$P(H_i / \theta) \approx K.\exp\left(-\frac{H_{yi}^2 + H_{xi}^2}{\Sigma^2}\right) \exp\left(-\frac{(H_{xi}R_{xi} + H_{yi}R_{yi})^2}{2\sigma_H^2}\right).$$

Vectors $R_i = T_i / ||T_i||$ and

$$T = T_i = \left(\frac{\partial x_s(t(i))}{\partial t}, \frac{\partial y_s(t(i))}{\partial t}\right)^T$$

are constant along an edge. $T_i$ components are also B-spline functions having spline basis as follows:

$$\frac{\partial B_{n,i}(t)}{\partial t} = \frac{n}{t_{i+n} - t_i} B_{i,n-1}(t) - \frac{n}{t_{i+n+1} - t_{i+1}} B_{i+1,n-1}(t).$$

If the $O_i$'s are conditionally independent, then the global likelihood becomes $P(O/v_s) = \prod_i P(O_i/v_s)$, and $\hat{\theta} = arg.\max_\theta \prod_i P(X_i / H_i, \theta) P(H_i / \theta)$, leading to the minimization of the quadratic energy function:

$$E(\theta) = (d - B\theta)^T W (d - B\theta).$$

$d = (x_{o1}, ..., x_{oM}, y_{o1}, ..., y_{oM}, 0, ..., 0)^T$ is a $(3M \times 1)$ vector, $W$ is a $(3M \times 3M)$ diagonal matrix with $W_{i,i} = W_i + W_{i+M,i+M} = 1/\sigma_{Xi}^2$, and $W_{i+2M,i+2M} = 1/(T^2\sigma^2 H)$.

$B$ is a $(3M \times 2N)$ matrix whose components are $B_{ij} = B_j(t(i))$. Currently, the points $i$ are uniformly distributed: $t(i) = i/M$. $B_x$ and $B_y$ are given by

$$B_{xij} = NH_{xi}\left(B_{i,2}(i/M) - B_{i+1,2}(i/M)\right), \text{ and } B_{yij} = B_{xij} = NH_{yj}\left(B_{j,2}(i/M) - B_{j+1,2}(i/M)\right).$$

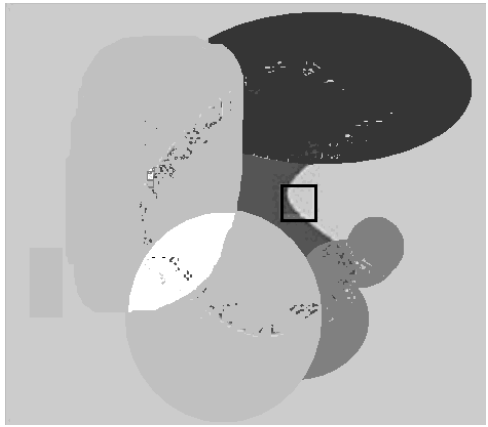

Fig. 2. Test image.

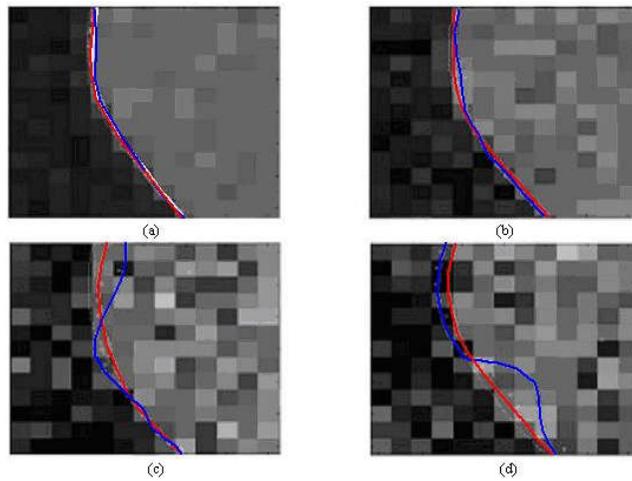

Fig. 3. Results for the suggested algorithm (red), the true edge (white) and the CSI (blue). (a) *10%* of noise, (b) *20%* of noise, (c) *40%* of noise and (d) *50%* of noise.

The solution $\hat{\theta}^* = arg\ min_\theta\ E(\theta)$ is given by the weighted least square relation: $\hat{\theta}^* = arg\ min_\theta\ E(\theta)$.

## IV. EXPERIMENTAL RESULTS

Several tests were performed with the squared area of a synthetic image (Fig. 2). For all the tests, the gradient was computed by means of the derivative of a Gaussian kernel with a standard deviation equal to *1*. The number of knots *N* was set to *M/4*.

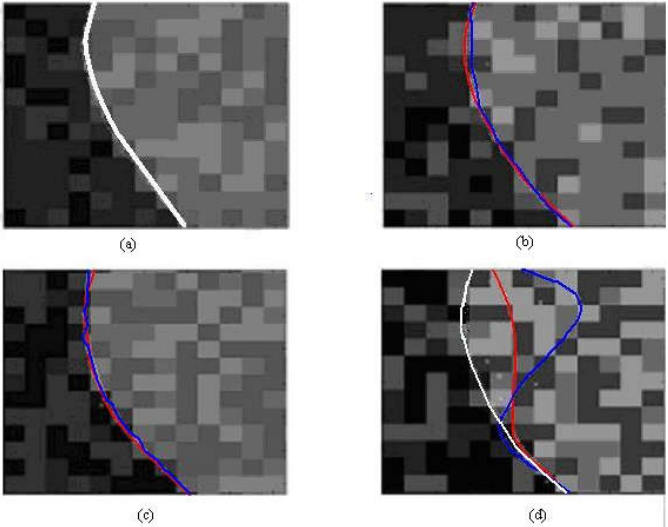

Fig. 4. Results for the suggested algorithm (red), the true edge (white) and the CSI (blue). (a) $p_0$ = 0.2, $\gamma$ = 0.3; (b) $p_0$ = 0.2, $\gamma$ = 0.6; (c) $p_0$ = 0.4, $\gamma$ = 0.3; and (d) $p_0$ = 0.4, $\gamma$ = 0.6.

First, this algorithm was tested for an additive normal noise with standard deviations equal to 0.1, 0.2, 0.4 and 0.5 which represent, respectively, *10%*, *20%*, *40%* and *50%* of the maximum amplitude. Fig. 3 shows the results obtained with this algorithm (red), the true edge (white) and the classical spline interpolation (CSI; blue). The suggested model has been developed for Gaussian noise, but it remains efficient for other types of noise as well. In Fig. 4, the results obtained with Salt & Pepper noise are shown, with $\sigma_n = \gamma \sqrt{2 p_0}$, for low rate noise ($p_0 = 20$) and high rate noise ($p_0 = 40$) with two different values of the amplitude $\gamma$. The Salt & Pepper noise pdf is

$$P(x) = p0\delta(x-\gamma) + (1-2.p0)\delta(x) + p0\delta(x+\gamma).$$

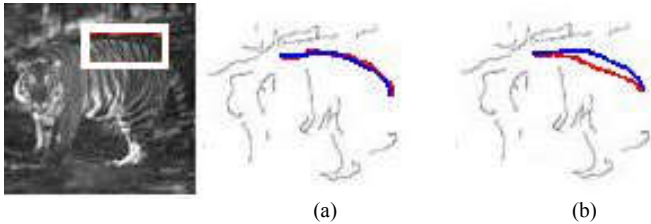

(a)          (b)

Fig. 5. Results for standard cubic B-spline fitting (red) and for stochastic cubic B-Sline (blue): (a) No noise; and (b) Gaussian noise with SNR=20 dB.

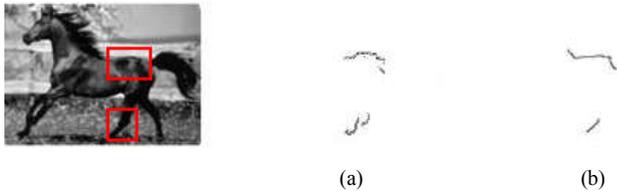

(a)          (b)

Fig. 6. Results for standard cubic B-spline fitting (a) and for stochastic cubic B-Sline (b) for Gaussian noise with SNR=20 dB.

Figures 5 and 6 show the performance of the proposed algorithm in real images. The rectangles mark regions where one can check the results of standard cubic B-spline and its stochastic counterpart for noisy images with SNR=20 dB. In Fig. 6, the performance of our algorithm seems to be slightly better.

## V. CONCLUSION

In this paper, a robust algorithm for edge estimation with sub-pixel accuracy based on a stochastic cubic B-spline is introduced. The model parameters are calculated according to an ML estimation framework relying on a Gaussian noise model. The data likelihood is computed from the observation model which includes both orientation and position information.

Comparative numerical experiments between this algorithm and the CSI have been carried out for two different types of noise: Gaussian and Salt & Pepper. The experiments have shown that our algorithm outperforms the other two, for both kinds of noise. It remains accurate even for high noise levels while the usual methods are generally sensitive to local perturbations due to the global computation of a given boundary using a ML rule. The likelihood of the observations is explicitly computed.

ACKNOWLEDGMENT

Prof. Estrela and Prof. de Assis are grateful for the support received by CAPES, FAPERJ and CNPq.